# BIM-assisted object recognition for the on-site autonomous robotic assembly of discrete structures

Mohamed Dawod[1] · Sean Hanna[1]

**Abstract**
Robots-operating autonomous assembly applications in an unstructured environment require precise methods to locate the building components on site. However, the current available object detection systems are not well-optimised for construction applications, due to the tedious setups incorporated for referencing an object to a system and inability to cope with the elements imperfections. In this paper, we propose a flexible object pose estimation framework to enable robots to autonomously handle building components on-site with an error tolerance to build a specific design target without the need to sort or label them. We implemented an object recognition approach that uses the virtual representation model of all the objects found in a BIM model to autonomously search for the best-matched objects in a scene. The design layout is used to guide the robot to grasp and manipulate the found elements to build the desired structure. We verify our proposed framework by testing it in an automatic discrete wall assembly workflow. Although the precision is not as expected, we analyse the possible reasons that might cause this imprecision, which paves the path for future improvements.

**Keywords** In-situ robotic construction · 3D object recognition · Pose estimation · Building Information Modelling (BIM) · Autonomous Assembly · Discrete Fabrication

## 1 Introduction

### 1.1 Context

Robotic automation stands as a promising field for construction, as it offers a safer and efficient approach for handling building resources on site which extends the limits for more creative and customizable architecture (Giftthaler et al. 2017). Although construction robotics display huge potential, its application in real construction projects is still limited, compared to the massive deployment of robots in industrial environments.

Unlike the case of the deterministic industrial environment, where robots blindly execute predefined instructions, the construction environment is vastly complex and unpredictable, due to the nature of sending the robots and the building components to the site. Thus, it makes it difficult for robots to adapt to the continually changing site without being an active member in the process. Therefore, equipping robots with advanced vision capabilities is a must for allowing them to observe, adapt and cope with the busy environment of the building site. Consequently, dealing with unique, irregular structures require robust object estimation approaches to enable on-site autonomous robotic assembly applications.

### 1.2 Problem

Current object pose estimation systems are not optimised for construction applications due to several factors. First, a process of preparation is normally required to precisely identify an object to a vision system, in which all known building components on a site are first scanned or labelled for accurate reference into a system (Furrer et al. 2017). On a large scale, this can become a very challenging and unrealistic process to achieve. Second, there is a difficulty in coping with elements' imperfections or vague specification: many systems manipulate building components as abstract entities (Feng et al. 2015) which expect a perfect model in an ideal

✉ Mohamed Dawod
mohamed.dawod.15@ucl.ac.uk

Sean Hanna
s.hanna@ucl.ac.uk

[1] Bartlett School of Architecture, UCL, London, UK

world. This criterion is often not the case in construction environments. These limitations reduce the flexibility and speed of systems in accommodating the significant creative demands of current building construction applications.

### 1.3 Aim

This paper investigates a flexible 3D object recognition approach to easily reference building components to a vision system and be able to deal with objects' imprecision. Using a virtual scanning process which takes the "ideal" representation model of all the objects found in the Building Information Modelling (BIM) model as the guide for finding all "best match objects" on site, without the need for physically scanning or labelling the objects to reference them. To examine to what degree a simple object recognition system can detect objects with inherent errors or imperfections given an ideal model?

Our overarching hypothesis is that using the available BIM data as the reference model will have the potential not only to locate highly matched known objects with a threshold to accommodate material deviations, but also flexibility to find the best match objects from an unknown pile.

To test the proposed approach, we used the case of constructing a man-made prefabricated structure made of many unique rigid elements as a case study. This context provides an ideal testing ground for this paper hypothesis, as the assembly of such structure would require a precise system which able to locate and handle such low tolerance structures.

Our goal was to detect and manipulate the best-matched known objects in an unsorted pile on site according to a given design in the form of the BIM model (see Sect. 3). Therefore, we developed a holistic approach that includes our detection method, and an automated process to grasp, manipulate, and determine the assembly sequence based on the design scheme of the structure and awareness of the surrounding context (see Sect. 5).

We assessed the system through a physical experiment and implementation of the automated assembly process of a discrete wall structure (see Sect. 6). As we were matching the physical objects to a virtual model, we were observing and analysing the level of tolerances, differences between them and the degree this could affect the resulting structure (see Sect. 7). Then, we discuss the overall performance of our method and its advantages, limitations, and its potential use cases (see Sect. 8).

### 1.4 Contribution

The main contribution of this paper is a BIM-Assisted Recognition approach which utilises the guidance of the recognition system by virtually scanning all objects in the digital design BIM model. This framework provides a simple method to reference objects to a vision system and tolerant system which can handle objects impression or vague specifications. This proposed recognition system is not only relevant to scenarios of the mass customization, where we have to deal with lots of highly matched and specified elements, but it could be also useful in the cases of dealing with unspecified structures to find the best-matched objects, like the case of using natural materials in building dry stone walls.

## 2 Related work

### 2.1 Object recognition in robotic construction

Many attempts tried to increase the level of automation in construction by facilitating the communication between the BIM information and real construction processes through proposing different methods of machine vision to enable the robot to handle the building materials on site. New procedures are required to replace the current complicated and time-consuming indexing and referencing methods used to facilitate the workers during construction (Krieg et al. 2015).

Feng et al. (2015) proposed a mono-vision marker-based system for referencing and rapidly localising non-unique and modular building components to automate the robotic assembly of free-form modular structures in a construction site. While it has some advantages in terms of fast localisation, it showed several feasibility, robustness, and durability limitations. However, the main problem with this method is that it deals with objects as abstract entities and using a referencing system to enable communication with the robotic system, with no notion of the actual model.

Our method tackles most of these limitations, as it is not limited to modular shapes or light condition, and more importantly, it is sensitive to material variances and shape changes. Moreover, it requires no need for any setup or contact with the physical model in the preparation phase to reference the objects, which makes it very generic and efficient on many levels.

Sandy and Buchli (2018) proposed a monocular-vision object-tracking system to retrieve a position of a camera to multiple known objects, using an edge-based tracking approach. This system was demonstrated in an augmented reality environment to guide a human builder to construct a complex brick structure by hand. However, it is currently not suitable to guide an autonomous robotic manipulator, as it is not able to search for and identify the objects automatically. Instead, it uses the object outline as a reference body to locate a camera to it and to achieve that it requires at the start orienting the camera to a specific point of view of the object for successful registration.

On the contrary, our system uses the knowledge of shapes to automatically search and locate the potential objects which match a target object from any point of view on the site and uses the knowledge of the design to assist the robot in manipulating the found objects.

An automated stone stacking process was presented by Furrer et al. (2017); this system uses a pose estimation process guided by a physical 3D scan models of a set of stones to locate them in a scene. Then, it uses a pose search algorithm to sequentially decide how to place them. We can think of it as playing the Tetris game (a Tile-matching video game), where the system takes a specific stone and looks how to fit it into the wall. However, in the real cases of building with dry stones, a skilled builder often specifically selects specific stones for particular areas as a sort of foresight planning (Villemus et al. 2007).

Our approach, instead, assumes that the assembly sequence exists in the form of a specific design BIM model. Besides, our method requires no physical scanning of the objects to reference them, as this would be a very tedious process in a real construction site. We replaced this step with the virtual scanning process, which takes the reference from the objects virtual model in the BIM domain, allowing the system to search for the best-fit objects on the site.

## 2.2 3D object detection

Vision sensors for object detection and pose estimation have been the focus of research within industrial robotics applications and in the development of the human-like robots, e.g., Willow Garage PR2 robot, as robotics to work in industrial or human environment have to deal with various types of object with different geometries which require a more generic approach to cope with such variety. Advances in 3D vision systems have led to a growing interest in object recognition and pose estimation techniques which operate on three-dimensional data. Furthermore, the knowledge of the 3D geometric shape and the pose of an object greatly facilitates the execution of a solid grasp.

Kuo et al. (2014) proposed the use of 3D object detection and pose estimation for fast industrial Robotic Bin Picking application to recognise an industrial object from a pile of the same object. Using local detection method which depends on finding local key points features on both the object and the scene point cloud. In contrast, in a construction context, we are expected to deal with invariably different types of geometries.

Papazov et al. (2012) presented a fast object recognition and pose estimation approach which use point cloud data of a Microsoft Kinect sensor to recognise multiple known household objects and perform robotic pick and place in unstructured environments. The object referencing in this approach was based on using the 3D-scanned models of the objects; in our case, we use the computer aided geometry (CAD) model provided by the BIM model.

The most relevant precedent is an approach proposed by Li et al. (2015) to complete the missing information in a 3D scan reconstruction scene, using an object recognition method that retrieves and replace found objects in a scene with the closest match in a 3D hand-modelled shape database. Our approach differs for its accuracy, context, and the use of 1:1 scale models, as the scale is a vital element in the assembly process.

Moreover, the BIM database used in our case not only serves as a shape guide, but also it offers the physical and mechanical specification about the models and its spatial relation to other elements in the design which helps significantly in guiding the robot to handle such structures.

Wong et al. (2017) proposed an object recognition and pose estimation approach utilising a trained Convolution Neural Network (CNN) to identify and segment objects of interest in a scene from an RGB image (Semantic Segmentation), and then, it applies this segmentation image as a mask to crop a 3D scene point cloud. Later, it uses an object library of 3D-scanned models of known objects to retrieve the 6DoF pose of the detected shapes using point cloud matching algorithms.

In this context, this approach relies heavily on massive data sets for training and to be able to identify and segment objects (i.e., the system was trained on 7500 labelled objects to detect automotive objects). Taking into account that it is targeting industrial applications, where the frequency of the product shape changes is quite low, therefore, it is acceptable to have such training process, unlike in an architectural context, where the frequency of shape change is relatively high and applying such training process for every new design model would be unrealistic. Therefore, our recognition method relies notably on extracting the unique geometrical features of each object's 3D point cloud.

In the above examples, even though object recognition and pose estimation are widely used in the industry and robotic community, its uses in a robotic architectural application is still quite limited. These approaches are challenging as objects typically need to be 3D scanned to be referenced into a system to offer more reliable detection.

Object recognition is commonly used in the bin-packing application. In this application, grasping is not an issue, as there is no predefined pose to how an object should be placed in a particular location. However, in the building construction case, the object has to be precisely grabbed and placed in a predetermined pose and location which is quite a challenging problem and require more advanced planning techniques.

Therefore, in this text, we propose the integration of the BIM model into the process to assist construction robots on site with two things: (1) virtual shape library to guide the robotic detection and (2) spatial design layout to facilitate the manipulation of the targets.

## 3 BIM-assisted object recognition

In this section, we describe our object recognition framework. In Sect. 3.1, we first give a brief overview of the object recognition and pose estimation problem, the primary methods used in this area and the chosen pipeline. In Sect. 3.2, we give an overview of the training process methods. In Sect. 3.3, we describe our virtual scanning process. In Sect. 3.4, we explain the implemented recognition and pose estimation process.

### 3.1 3D object recognition and pose estimation

In the field of computer vision, 3D object recognition is a method used to identify objects correctly in a point cloud. It is usually carried out with the term pose estimation to compute the six-degree-of-freedom (DOF) transformation of the recognised model. This method works by finding a set of correspondences between two sets of point clouds, one of them represents the object we are looking for, and the other represents a scene with potential candidates. These correspondences are defined based on extracting the significant features or the signature of a geometry called descriptors.

A descriptor is a method of encoding a point or a whole object point cloud with notable features in a simple form of a histogram which represent the object signature. It is used to accurately and rapidly identify an object across multiple point clouds without being affected by the quality of the point cloud. For example, a Viewpoint Feature Histogram (VFH) is based on computing the angles between the point cloud normal and the vector of the point cloud's centroid to the viewpoint component (Aldoma et al. 2012b).

In the literature, 3D object recognition can be approached from a variety of perspectives. Two major methods include local feature-based methods and global feature-based methods (Aldoma et al. 2012a). The local feature-based methods only extract the main key points of a 3D surface of a model and a scene, then associate the descriptor with its local neighbourhood, as they apply a descriptor matching process which goes through all the descriptors in both scene and models until it finds a reliable point correspondence. The main advantage of this method is its ability to recognise objects on clutter and occlusion scenes.

In contrast, global feature-based methods generalise one single descriptor for the whole object surface; this approach requires a 3D pre-segmentation process to localise the possible object instance in the scene. For each 3D segment extracted on the scene, it calculates the descriptors and searches for matching descriptors in the model database, yielding model hypotheses. Unlike the local approach, it cannot be directly applied to cluttered scenes. Although it is less effective in the presence of partial object occlusions, the global approach is characterised by a smaller complexity in the description and matching stage compared to local methods.

Even though the locally based system seems the appropriate choice for our process, defining local features depends heavily on local surface information that is directly related to the quality and resolution of the acquired model data (Alhamzi et al. 2015). As we are interested in utilising an affordable depth sensor like Kinect as our primary source to gain the point cloud and because of the nature of Kinect to provide noisy data, we cannot rely on such method. Therefore, the global feature-based approach was used as our main pipeline for its flexibility to work with noisy data. This method is also beneficial for its memory footprint, since a notably reduced amount of information needs to be stored to represent the model library (Mian et al. 2010).

### 3.2 Training process

A vision system is usually trained by taking several snapshots of every object from different angles and computing the desired descriptor for each snapshot, then storing them in a database. The training phase happens by either scanning the physical object from several points of views using a depth sensor and a calibration system or using the 3D scan model of the object and a rendering system that simulates the physical scan effect by placing several virtual cameras around the object to get all the desired viewpoints with no calibration setup (Aldoma et al. 2012a). However, both techniques require a very sophisticated setup which requires physical interaction with the model.

Creating a training data using real objects have the following limitations: (1) it is difficult to capture all the viewpoints and the poses of the objects, especially in our case dealing with architectural objects which could be very difficult for many reasons like scale, weight, time and cost. (2) It is difficult to place the object at the same pose of its virtual model, causing differences in the transformation, resulting in a false pose.

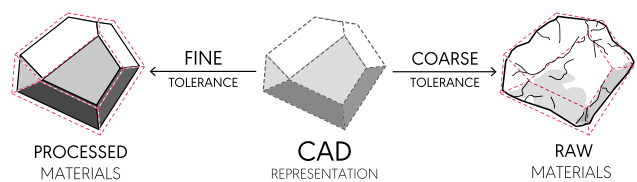

**Fig. 1** Range of the detection applications, finding the best match object to the desired object depends on the level of tolerance

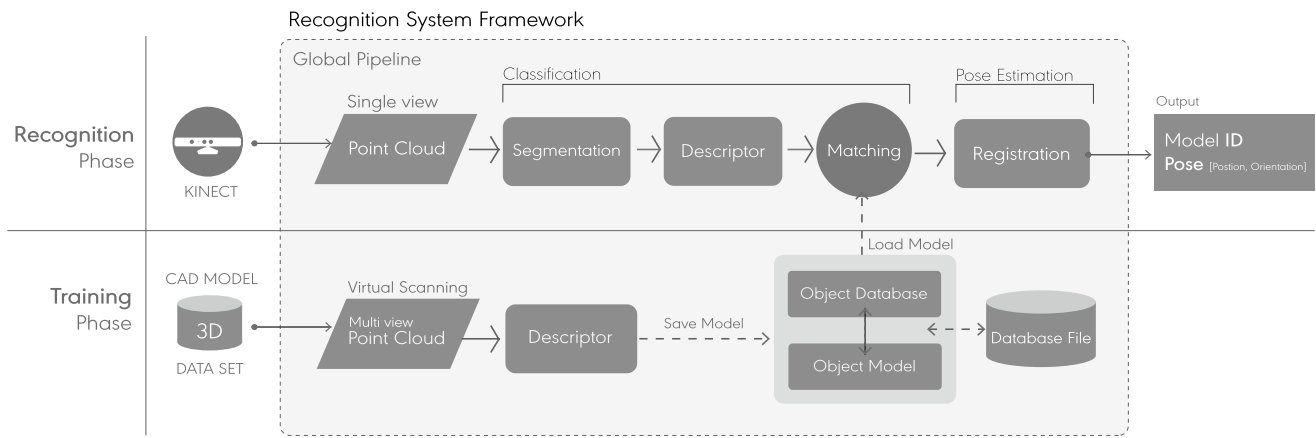

**Fig. 2** Recognition System Framework, which comprises two different phases: an online recognition phase and an offline virtual training stage

Therefore, it would be preferable to have an approach that uses the 3D CAD representation of the objects coming from the BIM model as the base model for training instead of using physical scan models of the objects. This method has the following advantages: (1) it would be affordable and require no preparation or setup processes and the actual model does not have to be present; (2) not limited to the scale of the object; (3) the object pose will be the same as the virtual model which will have a significant impact on the retrieving precise pose for assembly; and (4) the availability of the 3D representational model of the objects from the BIM model.

Besides, having a precise training model coming from the BIM model will not only allow the system to detect matched prefabricated objects, but also it can be flexible to search for unspecified but similar objects like natural found materials such as stones (Fig. 1). This adaptability is possible due to having this approach treat all the objects in a scene as a point cloud, which gives it the flexibility to treat all the structures in a scene equally as potential candidates. Even if the object does not perfectly match the target shape, it still can recognise it as the closest match if it meets a certain threshold (this is described more in detail in Sect. 3.4).

The recognition framework (Fig. 2) comprises two different phases: a virtual training stage and an online recognition phase, in which the real scene point cloud gained using RGB-D (Microsoft Kinect) sensor and processed using a standard global recognition pipeline as mentioned at Aldoma et al. (2012a) which involve a segmentation, recognition and poses estimation and final refinement of the recognition results stages. In the following text, we will provide more insight details about the recognition algorithm to provide the intuition on how it works, which will help us later to understand its potentials and limitations.

### 3.3 Virtual scanning

While pose estimation is about finding the transformation applied to a given object in the space from a specific frame of reference (usually at the object centroid), it is essential for assembly to know the needed transformation to place the detected object to its corresponding pose in the design. Therefore, both the training object and its equivalent object in the model have to share the same object frame of reference. While the virtual scanning process uses the object centroid as the object frame of reference, and the detection system always estimates the object pose in relevance to this reference. Therefore, the training objects have to be reoriented from their centroid coordinate frame in the design model to the world reference.

To address these criteria above, we perform a preparation stage before the virtual scanning, which operates as follows: assuming that the design model coordinates is the world zero coordinates: (1) we calculate the centroid of each object in the BIM model to represent the object coordinate frame in the design model. (2) We orient all the objects in the model from its new frame of reference to the world coordinate frame. (3) We keep all the objects frames in the design model in the memory to be used later in Sect. 5 to assist the assembly planner. (4) We save the oriented object and its ID to a database ready for the virtual scanning step (Fig. 3).

The virtual training is a process of simulating the scanning process using a depth sensor to gain a multi-view-point cloud snapshot of an object. Where for each object $O_i$; $i \in \{1, \ldots, n \text{ Objects}\}$, an array of virtual cameras are placed uniformly on a sphere with radius large enough to cover the object, and each camera acquires a partial point cloud view $P$ of the object by sampling the depth buffer of the graphics card. Afterwards, the system calculates the descriptor $D$ for each point cloud scan $O_{i,P,D}$; $D \in \{1, \ldots, n \text{ } P\}$, then it stores each object point cloud snapshot with its descriptor

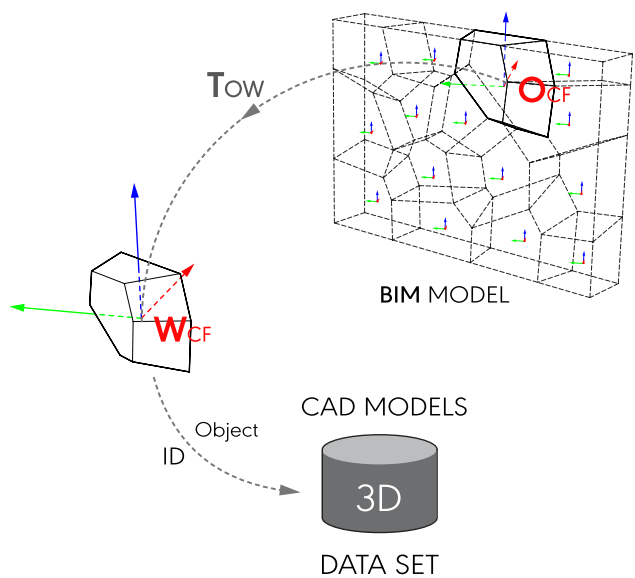

**Fig. 3** Data preparation process, where we orient all the objects in BIM model from its centroid coordinate frame $O_{i,CF}$ in the design to the world zero coordinate frame $W_{CF}$. Moreover, we keep all the objects frame of reference in the design model in the memory and save the oriented object and its ID to a database ready for the virtual scanning step

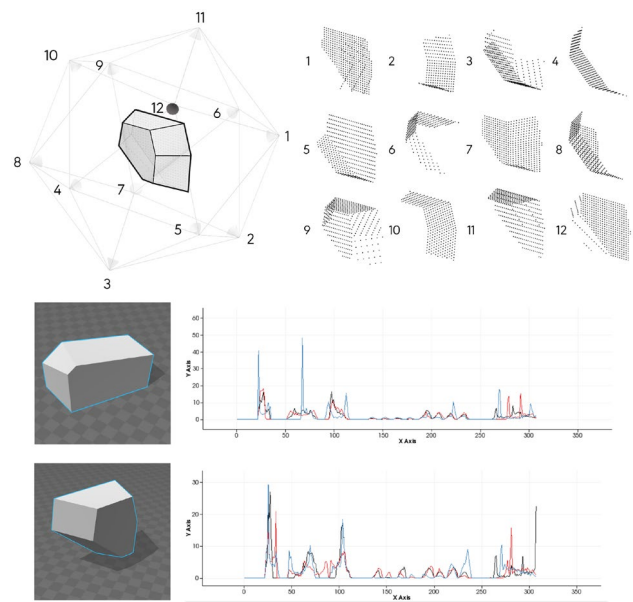

**Fig. 4** At the top, a visualisation of the multi-viewpoint cloud virtual scanning process used in the training process, where create a database of all the objects multi-viewpoint cloud snap shots and its respective calculated descriptor. At the bottom, a histogram example of a computed descriptor for one scanned faces of two different objects

to a database ready for the recognition stage (Aldoma et al. 2012a) (Fig. 4). In our case, we used the default setting of 80 virtual cameras and resolution of $150 \times 150$ for the synthetic depth images to train our system.

### 3.4 Recognition stage

At the recognition stage, we get the scene point cloud $S$ from the Kinect sensor. The process starts by segmenting the scene (a process of splitting a scene into various objects) using the dominant plane extraction (a process to segment all the plane surfaces from the point cloud, like floors or walls or ceiling and considers only the clusters on it). The point groups $S_{P'}; P' \in \{1, \ldots, n \text{ Groups}\}$ result of this step represent a potential object, to be recognised dependently, for each of these we apply the following: (1) compute the descriptors $(D')$ $S_{P',D'}$. (2) For each $S_{P',D'}$, find the $N$ closest descriptors in the training set $O_{i,P,D}$ using the nearest neighbour (NN). (3) Then, the best $N$ candidates are selected $N'$ by a given threshold. (4) For the resulting $N'$, the pose estimation transformation is calculated $T_{SO,j \text{ matched}}; j \in \{1, \ldots, N'\}$. (5) After aligning the views, a point-to-point Iterative Closest Point (ICP) (Rusu and Cousins 2011) step is used to improve the alignment $T_{SO,j \text{ refined}}$ (Aldoma et al. 2012b). (6) Select and return $T_{SO}$ with the highest ICP score. (7) In the case of low tolerance objects, we send the transformation $T_{SO}$, and the index of the selected object $O_i$ to the assembly planner (see Sect. 4) to retrieve the same model from the BIM model, or in the case of high tolerance objects we return the point cloud data $S_{P'}$ beside the transformation $T_{SO}$ and the index to facilitate the grasping.

## 4 Object to robot calibration

To place the object $O_i$ in the robot frame $R$, while the 3D camera is mounted on the hand of the robot, we need to perform a calibration process to know the relation of the camera to the robot. This problem is widely known as the hand–eye calibration problem. The hand–eye calibration is a process to calculate the position and rotation of a camera mounted on a robot, where the robot with the mounted camera is moved to multiple locations and in each one, the camera is calibrated by getting its location to a calibration body and from the location of robot-hand to the robot basis. In this part, we used this approach but with two important modifications.

(1) The calibration process was done manually using a 3D object which in this case was a box as the calibration body to locate the position of the camera $C$ roughly to the robot arm $T_{RC0}$ (Fig. 5). By moving the robot arm to three visible points on the object (the corners) $T_R = Pt_{R\ 1,2,3}$. Then, by scanning the same object with the camera mounted on the robot, we select the same corner point from the scan $T_S = Pt_{S\ 1,2,3}$. By solving the transformation difference between $T_X = T_R - T_S$ and applying $T_X$ to $T_{RC0}$ we locate the camera position to the robot arm $T_{RC}$. By having the robot camera transformation $T_{RC}$ and the transformation of the selected object with respect to the camera $T_{SO}$, we can easily solve the Robot Object transformation by $T_{RO}$; $T_{RO} = T_{SO} \cdot T_{RC}$ (Fig. 6).

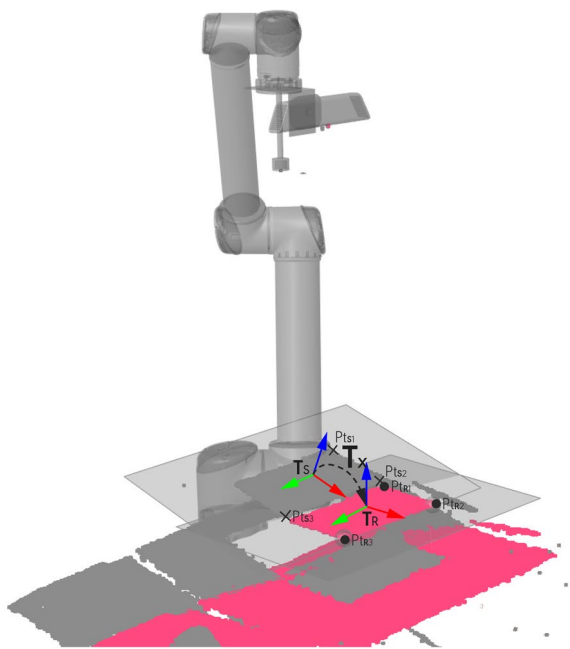

**Fig. 5** Manual calibration process used to solve the relation of a camera to the robot, where we scan a calibration body and measuring the transformation difference $T_X$ between the camera frame $T_S$ and the real-world frame $T_R$

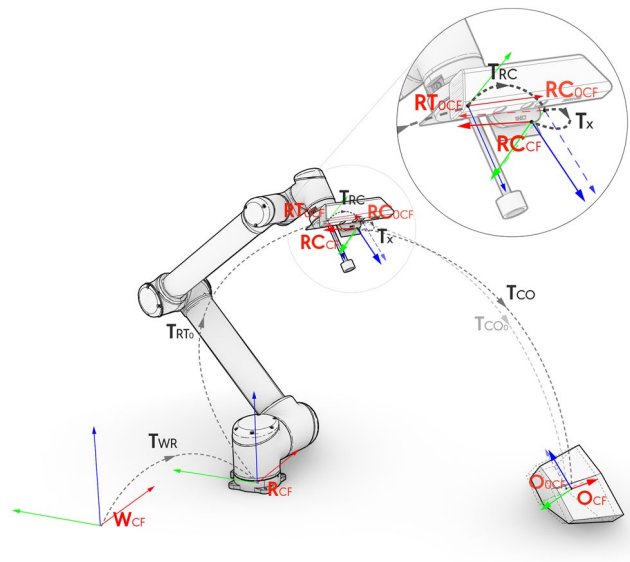

**Fig. 6** Coordinate frame setup to place the detected object $O_i$ in robotic frame $R_{CF}$. $W_{CF}$ denotes to the world coordinate frame, $R_{CF}$ is the robot coordinate frame in respect to $W_{CF}$, $RT_{0CF}$ is the tool 0 coordinate frame in respect to $R_{CF}$, $RC_{0CF}$ the initial camera coordinate frame in respect to $RT_{0CF}$, $RC_{CF}$ is the final camera position after applying the calibration data $T_X$, $O_{0CF}$ is the detected object coordinate frame in respect to $RC_{0CF}$, $O_{CF}$ is the final position of the object in respect to the final $RC_{CF}$ after calibration

(2) An additional step of calibration was added to the recognition data due to the transformation differences between the data coming from the recognition app and the scanning data used in the calibration. Using our detection app to recognise the same calibration object we then calculate the transformation difference between the scanned object and the recognised one, and apply this transformation to the data coming from the recognition server.

## 5 Automated assembly

To put our system into an application, we designed an autonomous construction workflow, where the robot is sent next construction site, and then, all BIM information of the design shapes and its spatial properties are feed into an Autonomous Robotic Assembly (ARA) system. Allowing it to automatically look for all the available resources at a site either processed (Prefabricated objects using Computer Aided Manufacturing CAM) or raw materials, and it autonomously generates and performs the assembly task to build the desired design.

The Framework of the ARA system comprises two steps, our recognition framework and an assembly planner process (Fig. 7). Both processes depend on the BIM information to work; first, the recognition system requires the BIM models of the different typologies the system will look for, and the second, the Assembly planner requires the actual design goal scheme to calculate how to manipulate the objects based on the intended design and awareness of context.

Building with irregular and discrete components is considered a difficult task, due to the high dimensional search space of grasping and manipulating an object to be placed in a specific location. Giving the object geometry and pose and the manipulator degrees of freedom and assuming that the assembly sequence was pre-assigned and provided by the BIM model.

Therefore, we developed a motion planning algorithm (Fig. 8) to search for the best grasping options and plan a collision-free motion to stack an object. This algorithm was based on the following parameters: (1) the pose and shape of the detected object, (2) its respective object shape and location in the design, (3) assembly sequence, (4) awareness of the context (the built environment and the already built structure), and (5) the robot position.

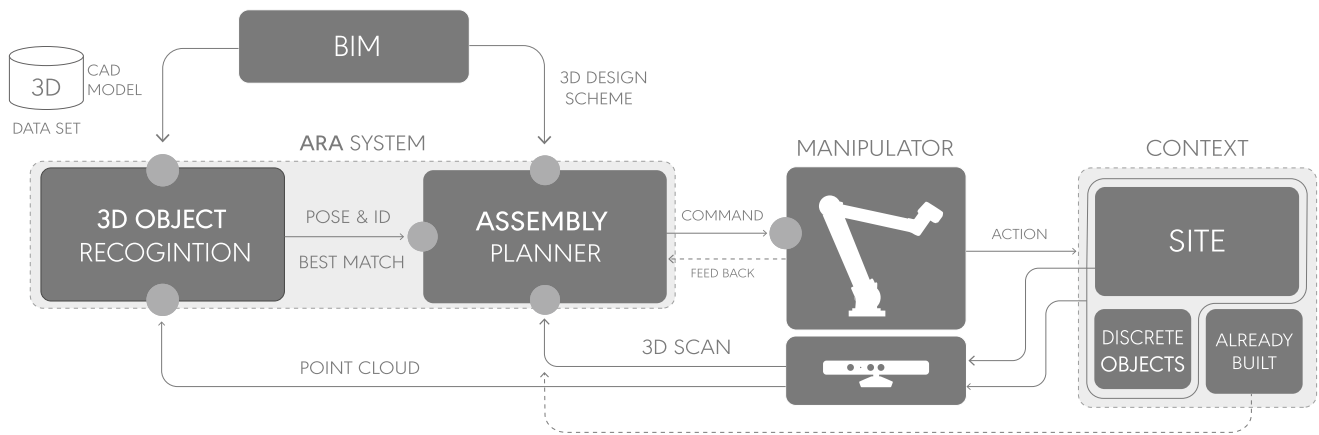

**Fig. 7** Autonomous robotic assembly system framework

Based on the building component sequence in the design, the assembly system waits for the vision application to locate the object and send its ID. By receiving a confirmation of detecting the required object ID, the assembly planner starts the following logic:

**(A) Grasp estimator**

**Inputs** (a) Tracked object (Object A) Pose, ID, Mesh $M_A$, its respective object from the design (Object B) pose, Mesh $M_B$. (b) The Context; 3D scan mesh of the Site $S$ and Already Built Structure (ABS). (c) The robot position $R$. **Output** (a) Two robot targets frames (grasp point) $T_1, T_2$.

**(A-1) Generate grasp points samples** (1) Using Object B circumference sphere radius, we create a Geodesic Dome around it to evenly project points on the available areas on the geometry. (2) In respect to the Object mesh, we calculate the normal of these points, create a range of possible reference planes $G_b$. (3) Orient the generated samples $G_b$ to Object A $G_a$ (Fig. 8, 1). (4) Using the normal of both sample frames $G_a, G_b$, we exclude all the points which collide with the context.

**(A-2) Search for the best grasp options** After filtering the sample points, we start our search by checking if the resulting points could lead to the assembly of objects or not.

(5) We virtually move the Object B in all the possible direction of all the remaining samples and use collision detection, we then exclude all the units which could lead to a collision with the context (already built, environment) (Fig. 8, 2). If any samples are left from the previous step, we start a validation process to test and filter these remaining points. By feeding these points as targets, we simulate the robot behaviour for each of these points on both objects. (6) Exclude all the targets in both list if any of them could cause errors in the robot control (e.g., Out of reach targets). (7) If any targets remained from the previous step, we use collision detection to exclude all the frame targets which would lead the robot to collide with the context (Fig. 8, 3).

After the last step, if any targets remain this means that the system was successful to find a grasp point option to pick up the object. However, in some cases, the system could discover several possible options and as our system needs only one. We select the targets based on the smallest distance to the robot from both object A, B (Fig. 8, 4).

**(B) Path planning**

Giving the two targets from the previous step, we generate a working path by creating a simple curve by interpolating the two target position and normal as a tangent (Fig. 8, 5).

For the robot to perform the assembly, we generate a list of sequential commands as follows: (1) Go to $P_a$ Target above Target A, then move slowly to it. (2) Gripper on. (2) Move back to $P_a$ with slow speed. (3) Move along the generated path, we use plane interpolation between the two targets to guarantee a smooth transition. (4) Go to $P_b$ target above Target B and move slowly to it. (5) Gripper off. (6) Move back to $P_b$. (7) Go to home position.

While path planning is not the focus of this paper, it is worth mentioning that even though this ad-hoc method guarantees a feasibly start and end joint pose. This method does not always ensure either collision or kinematic feasibility on the targets around this generated working path.

## 6 Case study: discrete wall assembly

### 6.1 Experiment setup

To measure the effectiveness of the presented recognition method, we tested the system by building a complex prefabricated structure made of multiple high tolerance objects with very distinct typologies. We assessed the performance in the three main areas: (1) the ability to identify and classify the objects; (2) the accuracy of the pose estimation the level

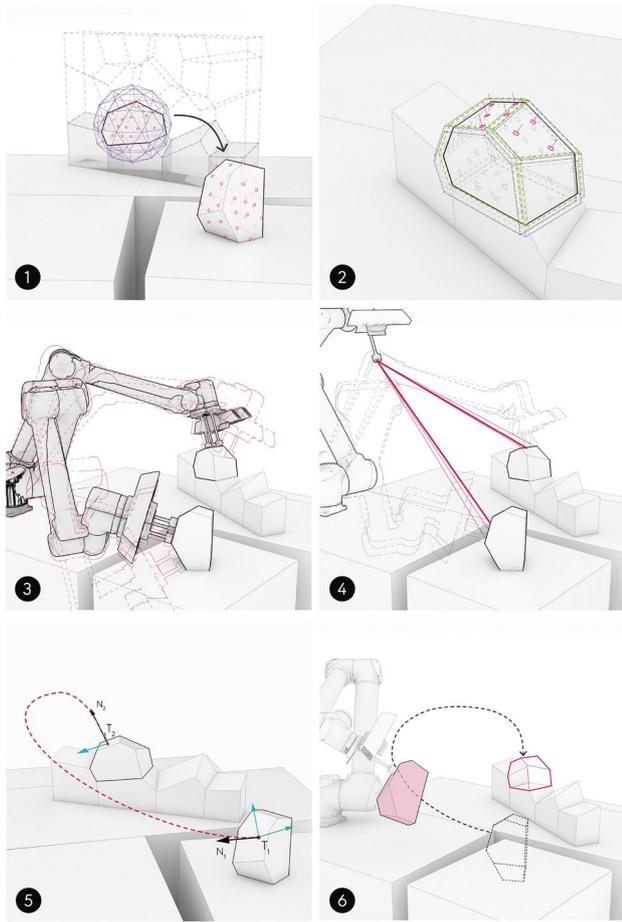

**Fig. 8** Automated grasp and path planner; searching to find the best grasping option to pick up the object and planning the movement concerning the context and the object pose in the design. (1) Using a geodesic dome we generate sample point around the object, exclude the points collided with the context. (2) Excluding all the target samples could cause the object to collide with the context. (3) Excluding all the target could cause the robot to an internal error in the robot control or collision with the environment. (4) Selecting between the grasp options available based on the shortest distance to the robot from both objects. (5) Generating the working path based on the target locations and orientation. (6) Execution of the assembly instruction, simulation of the robot performing the task

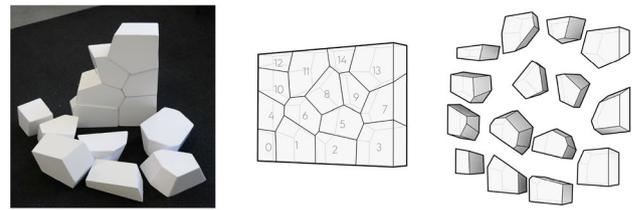

**Fig. 9** Left, the fabricated objects—middle, the virtual design of the whole wall—right, the different typologies models

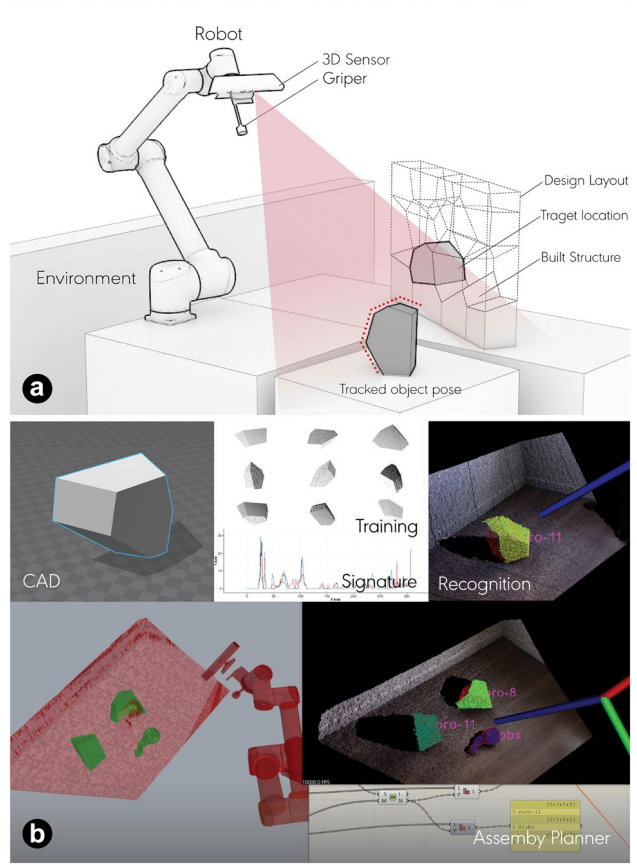

**Fig. 10 a** Experiment setup, **b** recognition app on top, the grasshopper vision and assembly planner in bottom of tolerance between the physical and the virtual model; and (3) the manipulation and grasp of the object using our assembly planner.

As the scope of this paper is geometry only, we designed and fabricated a structurally stable 3D Voronoi cell wall, Fig. 9), with the dimension of $800 \times 550 \times 120$ mm, comprise 15 blocks with different shapes and sizes made of foam core board. The Voronoi structure was used because of (1) the ability to provide different, unique shapes that form a clear structure and (2) the planar faceted cells of the Voronoi were easy to fabricate and offering a precise output, besides, and its suitability to grasp using a suction cup gripper. Besides, during the assembly of these models' faces, we intentionally apply various degrees of error at the joining, to give them some tolerance from the design ideal model.

We implemented the system in a Windows environment using the Point Cloud Library (PCL)[1] as the primary platform for developing our system. We used PCL to create a recognition app. In addition, we developed two plug-ins for grasshopper (Fig. 10b): (1) vision client plugin and (2) automated assembly planner. The vision plugin has three functions: (1) prepare the geometry for the training stage by saving the object automatically to the computer disk with its ID as the file name and in .ply extension; (2) calibrate the pose estimator to the robot; and (3) listen, decode and correct the transformation data coming from the recognition app (converting the pose data coming from PCL's right-handed coordinate system to Rhino's left-handed coordinate system). We used a 6-axis Universal-Robot UR10, an i7 CPU processor laptop with 16G Ram running Windows 10, the Robots[2] plugin controller to control the robot, a suction cup gripper and a Microsoft Kinect for Xbox depth camera mounted on the robot arm (Fig. 10a).

---
[1] http://pointclouds.org/.

In our experiment, we started by measuring the ability of the system to classify the object at any configuration, by placing the object in a specific location 1 m away from the Kinect to make sure that the camera can perceive the whole object. Due to the global recognition pipeline, which requires a segmentation process, we had to place the object on a flat surface. Then we present a set of eight angles every 45° to rotate our object and an upside-down pose. For each pose, we measure every 100 iterations, how many times the object was predicted correctly. In this test, we used the OUR-CVFH descriptor (Aldoma et al. 2012b) as the training method.

As the accuracy of the pose estimation affects the manipulation of the object which in turn could affect the final results, we ran the recognition system on several objects to measure the accuracy of the pose estimation and the level of tolerance between the virtual and physical objects. By selecting a point on the virtual object face (in this case, it was the face centre point), we are able to send this location as a target of the robot. Then, on the physical model, we check if the robot reached this target. Then we measure the error, by detecting the location the robot reached and the actual location the robot should have reached on the object. By sending the robot manually to the target point on the actual model, then we retrieve the manipulator position using the feedback component in grasshopper.

We tested the whole system by constructing the wall structure, to examine, to what degree does the accuracy of the recognition system affects the accuracy of the overall wall. This process happened in a continuous loop between the detection and assembly planner systems.

## 7 Results

The first experiment showed that the type of geometry and the viewpoint effect the results of the classification of the objects. Due to the high similarities between the geometry

---
[2] https://github.com/visose/Robots.

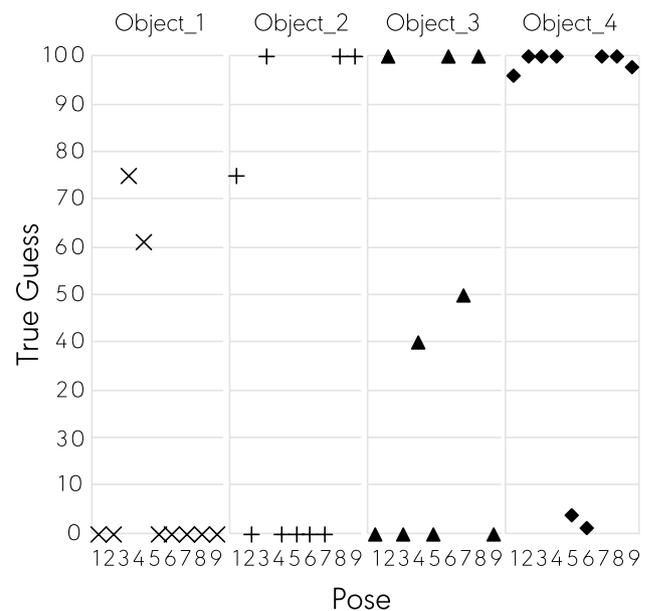

**Fig. 11** Object recognition and classification result for four typologies. Where the Pose of the object on the *x*-axis and the number of true predictions on the *y*-axis

of the objects, the recognition was inconsistent in some areas as it would get confused with other objects due to the close similarities between these objects from a particular viewpoint.

The graph (Fig. 11) shows the readings of four blocks, where the object different poses angle were represented on the *x*-axis as follow 1, 2, 3, … , 9 and each number represent an angular position, as discussed in Sect. 6, and the number of True predictions on the *y*-axis. The (Object-4) show the highest success rate seven out of nine and (Object-1) show the lowest success rate two out of nine poses, as it was likely to be guessed as with another object in the data set.

In the second experiment, tolerance difference in distances of 25 targets, it shows the difference between real and the virtual model result of the recognition. The graph (Fig. 12) illustrates the tolerance difference in distances of 25 targets were the targets represented on the *x*-axis and the tolerance difference in mm on the *y*-axis. It shows the difference between real and the virtual model result of the recognition with tolerance error in the pose between − 21 to 42 mm in *x*, − 9 to 29 mm in *y*, − 22 to 3 mm in *z*, which affect greatly in the grasping of the objects.

The process of building the structure in the third experiment exposed how the error factors can accumulate and affect the quality of the resulted wall. The system was able to grasp and manipulate most of the objects to build the wall (Fig. 13). It was successful until the fifth object, and the structure collapsed after placing the sixth object. Due to the tolerances found, causing the object to be shifted from the original design, in addition to other external factors; During the grasp or the placement, the end effector accidentally moves the object due to the light weight of the object, not having accurate feedback and assuring accurate data. Not having an adaptable design model based on the feedback, this caused the object to shift or collide with other objects.

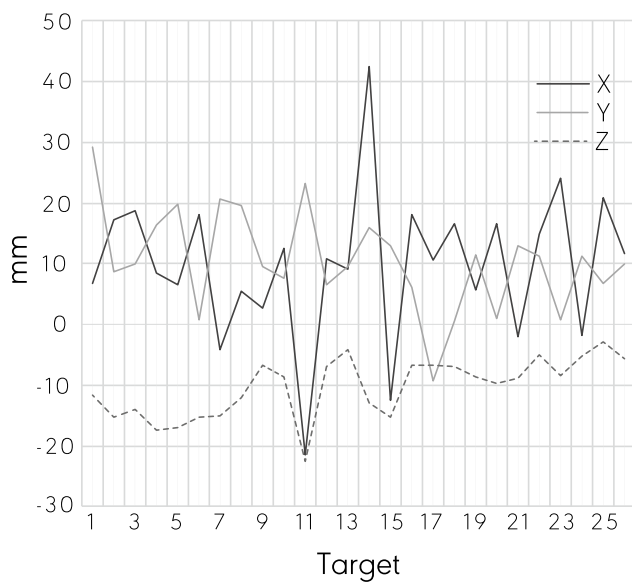

**Fig. 12** Accuracy of the pose estimation. Where *x*-axis represent 25 virtual and real targets, and the *y*-axis represents the distance tolerance difference between these targets in mm

In some cases, we had to reposition or rotate the objects manually to solve unsuccessful recognition attempts, or the inability to find a grasp or an assembly solution to manipulate the object because of the position of the object in respect to the manipulator. Repeating the test several times, in each time the object was grasped from in different poses producing walls with different qualities.

### 7.1 Limitations

In the case of building with low tolerance materials, we found the following limitations in regards to recognition:

(1) Surface details: objects with lower surface details (e.g., faceted geometries with large planar surfaces) is less successful in being recognisable. This issue can be traced back to the level of features extracted from the objects and the scene.

(2) Similarity: although some shapes may look unique, they may look similar to the recognition system as they might share a lot of similarity in shapes from a particular point of view, which affect the performance of the system in classifying the objects.

(3) Tolerances: the tolerance between the physical and virtual CAD model accuracy or/and the noisy data of the depth camera affects the pose estimation of the objects leading to false manipulation affecting the assembly robustness.

(4) Scale: objects need to be covered by the Sensor camera field of view. With sensors like Kinect, it cannot recognise a smaller object less than 5 cm, due to the data loss of the segmentation process. Partially covered objects which are the case of large shapes, the object level of details and uniqueness play a considerable role to find the object.

(5) Feedback: it is hard to recognise the object after placing the objects as they lose a lot of visible features.

(6) Descriptor: the method used to extract the geometrical features from the point cloud data of the models during the training on the detection process also affect the speed and success of the recognition and pose estimation system.

### 7.2 Critical assessment

It is worthwhile to mention that while the main focus of this research was about enabling the robot to locate and manipulate a 3D object autonomously. There were some constraints, we encountered during the process, which had a significant impact on the success of the process like the type of material, texture, weight and joinery. While these areas of inquiry were not a point of interest at this stage, not considering such factors would have drawbacks on the deployment of such autonomous systems in a real construction environment.

However, having the BIM model of a project as the main base to drive the autonomous process, in this case, would be beneficial as such information about the building components would be easily provided and found in a typical BIM model. Taking such information into account in future experiments by adding more layers of information to an autonomous process, would help it to plan and execute on-site efficiently.

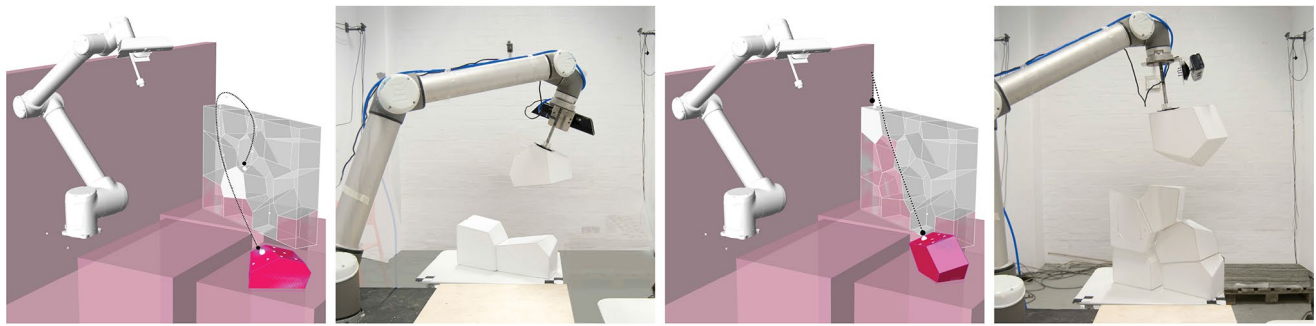

**Fig. 13** Assembly of the wall structure

Additionally, most of the problems and limitations related to the recognition part could be traced back to different factors such as the noisy data from the Kinect, the choice of descriptors, and the recognition pipeline. Each of these factors has a huge impact on the performance of a vision system.

## 8 Conclusion

The recognition system proposed showed that the method of using the virtual representation coming from the BIM model instead of having to scan the actual model brings advantages regarding flexible setup and affordable recognition. In addition to opening the possibilities for the system to be beneficial in either the case of prefabricated high tolerance assemblies or flexible assemblies of on-site natural or found materials.

The experiments showed the system able to detect and construct several structures with inherited imperfections within acceptable tolerances. However, it also showed several limitations related to the object's geometrical characteristic and its implication on the successful detection of the whole process success.

It showed that the object level of details and similarity of shapes from a particular point of view affects the recognition performance. The difference between the physical and ideal BIM model and noisy data of the depth sensor affects the pose estimation and consequently effects the assembly robustness. In addition to how the choice of the recognition pipeline affects the overall performance.

Despite the limitations, the proposed system showed great potential to be used as an efficient and sustainable approach for managing construction site resources and encourages human–robot collaboration. Moreover, it opens the door for the fabrication exploration of using or processing best-matched raw materials such as dry stones to achieve the specific design and the as-built documentation of complex assemblies.